\documentclass{article}
\usepackage[table,xcdraw]{xcolor}
\usepackage{bm}
\usepackage{fancyhdr}
\usepackage{ifthen}
\usepackage{textcomp}
\usepackage{setspace}
\usepackage{lineno}
\usepackage{cite}
\usepackage{booktabs}
\usepackage{array}
\usepackage{textcomp}
\usepackage[table]{xcolor}
\usepackage{amsmath}
\usepackage{multirow}
\usepackage[pdftex]{graphicx}
\usepackage[caption=false,font=footnotesize]{subfig}

\graphicspath{imgs}

\onecolumn
\begin{document}

\begingroup
\setstretch{1.0}

\title{Optical Ocean Recipes: Creating Realistic Datasets to Facilitate Underwater Vision Research}
\author{Patricia Schöntag, David Nakath, Judith Fischer, Rüdiger Röttgers, Kevin Köser}
\date{}

\maketitle

\begin{abstract}
The development and evaluation of machine vision in underwater environments remains challenging, often relying on trial-and-error-based testing tailored to specific applications. This is partly due to the lack of controlled, ground-truthed testing environments that account for the optical challenges, such as color distortion from spectrally variant light attenuation, reduced contrast and blur from backscatter and volume scattering, and dynamic light patterns from natural or artificial illumination.
Additionally, the appearance of ocean water in images varies significantly across regions, depths, and seasons. However, most machine vision evaluations are conducted under specific optical water types and imaging conditions, therefore often lack generalizability. Exhaustive testing across diverse open-water scenarios is technically impractical.
To address this, we introduce the \textit{Optical Ocean Recipes}, a framework for creating realistic datasets under controlled underwater conditions. Unlike synthetic or open-water data, these recipes, using calibrated color and scattering additives, enable repeatable and controlled testing of the impact of water composition on image appearance. Hence, this provides a unique framework for analyzing machine vision in realistic, yet controlled underwater scenarios. 
The controlled environment enables the creation of ground-truth data for a range of vision tasks, including water parameter estimation, image restoration, segmentation, visual SLAM, and underwater image synthesis. We provide a demonstration dataset generated using the Optical Ocean Recipes and briefly demonstrate the use of our system for two underwater vision tasks. The dataset and evaluation code will be made available.

\end{abstract}

\endgroup

\section{Introduction}
\label{sec:intro}
When engineering automated systems for ocean mapping, monitoring, or manipulation, cameras and machine vision offer promising capabilities. These technologies enable object and place recognition, habitat mapping and tracking, platform position stabilization, and drift correction.
However, the underwater environment poses significant challenges for visual sensing: local water properties and lighting conditions can drastically affect visibility and limit the effectiveness of machine vision. 
Water is a medium that has higher physical density than air resulting in greater impact of light-medium interactions on its optical appearance. Consequently, submerged objects often appear dramatically different from the same object photographed in air \cite{mobley1994light}. The main effects are due to higher light attenuation and more scattering of light. Higher attenuation reduces visibility and availability of sunlight illumination in deeper waters. Artificial light becomes necessary in that case but often adds dominant, co-moving light patterns to images \cite{song2022survey}. The spectrally variant characteristic of light attenuation additionally distorts color information. 
Scattering events between light and matter (water, dissolved or suspended (an-)organic particles, sediments) are macroscopically perceived as blur, due to forward-scattering and white haze, due to back-scattering. The presence of those effects can be highly variable between different regions and seasons of the ocean.

\begin{figure}[t!]
    \centering
    \includegraphics[width=0.7\linewidth]{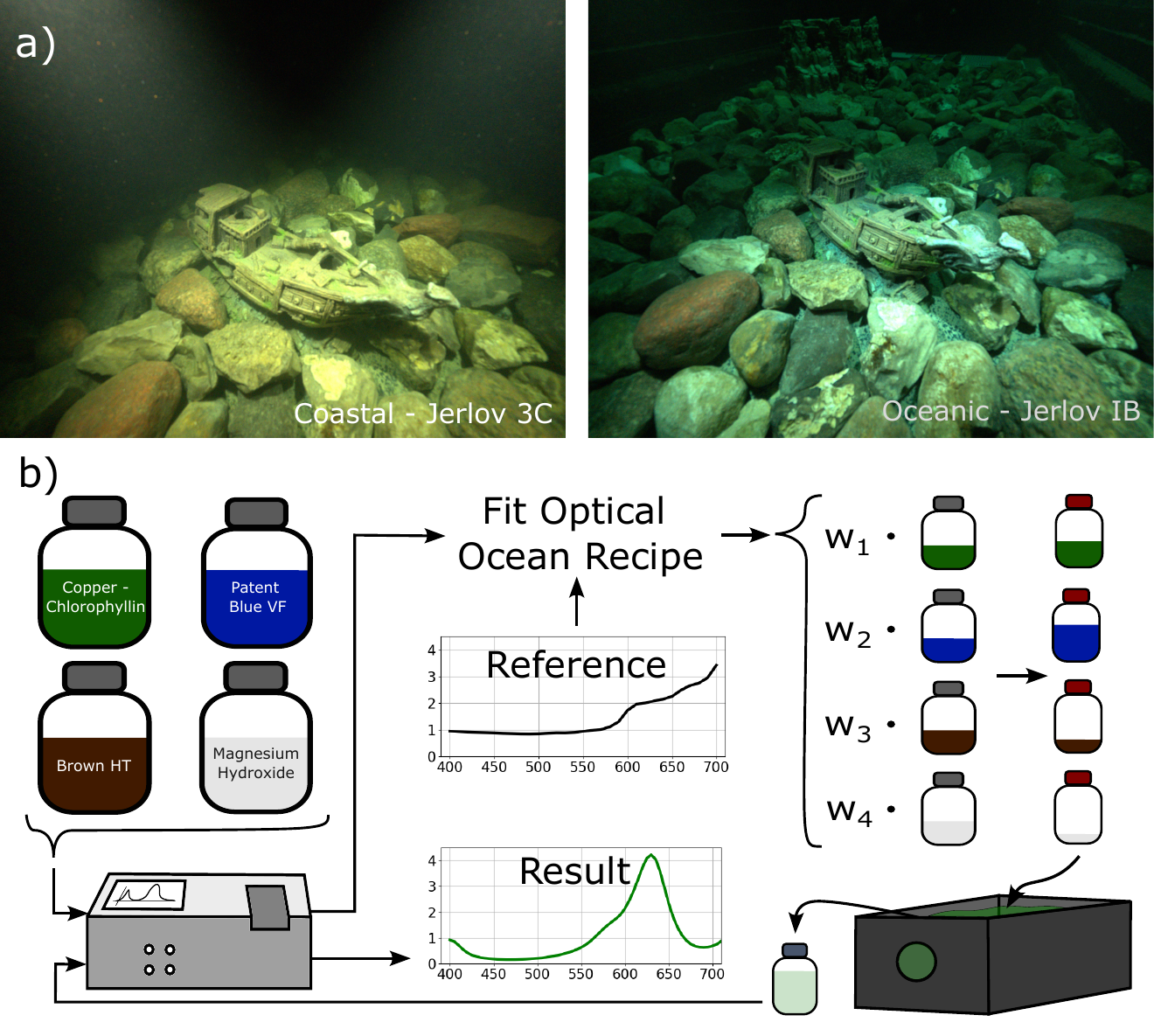}
    \caption{\textbf{Optical Ocean Recipes Method and Image Examples.} a) Images of same scene in a tank in two different water types. b) The \textit{Optical Ocean Recipes} are derived by fitting known optical additives to a reference spectrum. Using a spectrophotometer, the resulting colors can be accurately characterized and controlled. While our work focuses on the four specific additives and water references from \cite{williamson2022measured}, the Optical Ocean Recipes are not limited to these water references or to this particular number and type of additives.}
    \label{fig:Water_SetUp}
\end{figure}

Consequently, how state-of-the-art computer vision methods perform on particular ocean imagery is often only discovered during time-consuming trial-and-error experiments. Often, prior to ocean application, experience for a certain kind of task is limited to in-air performance tests. Whether or when performance evaluations on terrestrial datasets can be transferred to the subsea domain remains unclear. To investigate how computer vision methods can cope with the optical characteristics in an ocean region of interest, it is necessary to perform experiments on datasets that cover all the effects that may occur in a controlled and quantifiable way.
However, systematic performance analyses during concept design or evaluations in early prototyping phases of oceanic vision systems are hindered by a lack of adequate evaluation data, because ground truth is typically missing for real-world ocean scenarios: the appearance of the seafloor without water remains unknown, and the ocean’s optical properties, along with prevailing lighting conditions, are typically not recorded in image surveys due to the significant technical effort required. 
Additionally, method development or adaptation during a cruise in the ocean is tedious and comes with huge overheads and preparation, requiring an intermediate development step with easier access that is still realistic.
Making reasonable algorithm choices is even harder for real-time applications or integration into more complex general-purpose software packages where the specific challenges of a use case are unknown beforehand. 

To deal with these challenges of underwater vision research, we propose using \textit{Optical Ocean Recipes} to mimic the optical properties of real ocean waters in a test tank scenario by mixing defined amounts of optical additives into clear tank water. The tank environment enables the generation of controlled underwater images with ground truth for state-of-the-art evaluation protocols including, for example, precise camera poses or image appearance without the effects of water. 
Additional ground truth of optical water properties enables linking performance results to specific regions of the ocean, thus informing algorithm selection for specific use cases. 
A typical disadvantage of taking images in a tank is the limited scene depth. Achieving scene depths greater than one or two meters, closer to those found in realistic image surveys, would require a significantly larger pool, which is often impractical. Our ocean recipes offer a solution to overcome this limitation. We can increase the concentration of optical additives in the tank by a scaling factor and achieve the same image appearance as if more water is between the camera and objects.  
 
Synthesizing realistic ocean water with the proposed \textit{Optical Ocean Recipes} facilitates systematic testing and evaluation of underwater vision methods. The recipes allow researchers to align the optics of tank water with reference absorption and scattering spectra. We suggest using oceanographic measurements of Jerlov water types \cite{jerlov1977classification, solonenko2015inherent, williamson2022measured} to create realistic seawater conditions, while our method also supports the use of alternative references, such as measurements from specific regions of interest \cite{bi23modellingnaturalwaters}. 

Previous work on simulating visual effects of ocean water often left open the question of how closely the generated images match real-world conditions, and thus how transferable the evaluation results are to real use cases. In this work, we propose using a mix of food colorants whose spectral properties we measure in detail to generate realistic appearance in a controlled test environment. We also compare them to previously used optical additives such as milk and tea. Finally, we present a demonstration dataset generated with our Optical Ocean Recipes and show how it supports the application of state-of-the-art evaluation protocols. Our demonstration focuses on image matching and semantic segmentation, but the ocean recipes are equally suitable for generating datasets to evaluate other underwater vision methods, such as color restoration, visual simultaneous localization and mapping (VSLAM), or generative techniques. In this way, the Optical Ocean Recipes provide a practical foundation for developing and debugging robust computer vision approaches for the subsea domain.

\section{Related Work}
In this work, we aim to create image data that enables easy, controlled, and reproducible evaluation of the real-world performance and limitations of computer vision methods. A key requirement for such data is the availability of ground truth. Additionally, information about the water conditions is essential to link performance evaluations to real-world use cases. The first part of this section summarizes how this requirement was addressed in previous work.

The proposed ocean recipes provide precise control over water properties and lighting conditions, enabling more systematic evaluation of computer vision methods in underwater environments. This controlled setup not only facilitates the use of established evaluation protocols but also supports detailed analysis of performance limits under varying visual conditions. We demonstrate this on two representative tasks: image matching and semantic segmentation. The second part of this section focuses on earlier evaluation strategies and datasets.

\subsection{Creating Evaluation Data}
Prior research has addressed this by constructing evaluation datasets in water tanks infused with scattering or coloring agents \cite{codevilla2004achieving, garcia2011detection, yang2023knowledge, duarte2016dataset, jarina2021simulated, grimaldi2023investigation}. For example, the TURBID dataset \cite{duarte2016dataset} employs milk and, in smaller subsets, blue ink, green tea, and water-soluble chlorophyll, to mimic scattering or attenuation effects. Ground truth is provided in the form of a corresponding clear-water version of each image. For subsets created using milk, the mean squared error (MSE) between turbid and clear images is reported as a turbidity metric. Yang et al. \cite{yang2023knowledge} introduced the UFEN dataset for evaluating underwater feature matching, with ground truth obtained from controlled tank experiments. Talcum powder is used as a scattering agent, and the dataset reports both nephelometric turbidity units (NTU) \cite{kitchener2017review} to characterize the water and the structural degradation index (SDI) \cite{garcia2011detection} of the corresponding images. 

Related approaches in the field of optical communication \cite{cochenour2014suppression,mullen2011investigation} have used reference mixtures of known scattering and absorbing agents to reproduce optical water properties for laser communication studies. However, the focus in this field is on single wavelengths of certain lasers and therefore do not account for properties across the full visible spectrum

An alternative approach to generating evaluation data with accurate ground truth is synthetic data. Previous studies have created synthetic data either by superimposing the simulated effects of water onto images with existing ground truth \cite{song2022virtually, li2017watergan}, or by generating underwater images entirely using physically based rendering \cite{schontag2022towards,zwilgmeyer2021varos, nakath2022optical, sedlazeck2011simulating} or generative neural networks \cite{yang2025knowledge, li2017watergan}. Nevertheless, it remains an open question how to objectively quantify the gap between simulations and real underwater images. The performance of tested methods on synthetic datasets may therefore not directly transfer to real-world data.

Our proposed method enables an objective comparison to the created artificial ocean water to real-world ocean water. Additionally, adjusting the amount of optical additives to the targeted water properties adds more control and repeatability, and enables more diversity in evaluation datasets compared to previous tank-based setups.

\subsection{Underwater Robustness Evaluations}

Underwater image matching has typically been evaluated using indirect strategies that circumvent the lack of ground truth, such as (1) RANSAC inlier counts \cite{yang2025knowledge, shkurti2011feature}; (2) pseudo ground-truth from more accurate methods \cite{yang2023knowledge, garcia2011detection}; or (3) downstream metrics like pose estimation or SLAM \cite{boittiaux2023eiffel, jung2023performance, chen2023evaluation, aulinas2011feature}. However, RANSAC metrics can be misleading when structured lighting and weak visual cues in deep-sea images lead to consistently incorrect matches that still appear valid.

Yang et al. \cite{yang2023knowledge} and Garcia et al. \cite{garcia2011detection} constructed pseudo ground-truth datasets consisting of either keypoint detections or established correspondences on underwater images acquired in controlled tank environments with manipulated water properties. They evaluated the performance of either keypoint detectors in isolation or full feature matching pipelines, respectively, using clear-water versions of the test images. This evaluation strategy assumes the existence of an optimal set of keypoints that all methods should detect to perform well. However, methods based on fundamentally different principles of keypoint detection, description, or matching may be systematically disadvantaged under such a metric.

Earlier work acquired ground-truth poses for feature matching evaluation via pose estimation using different approaches depending on the image source. Joshi et al. \cite{joshi2019experimental} conducted experiments in the sea using external sensors such as Doppler Velocity Loggers (DVL) and Inertial Measurement Units (IMU). Other studies \cite{nielsen2019evaluation, grimaldi2023investigation}, including ours, utilize controlled tank environments, which enable more complex and precise camera tracking systems. Both approaches have advantages and limitations: navigational sensors (IMU, DVL) offer mobility but suffer from limited localization accuracy, whereas external camera tracking systems provide higher precision at the cost of requiring elaborate technical setups that may restrict experimental flexibility. In our work, we combine the strengths of feature matching evaluation through pose estimation with the concept of calculating pseudo ground-truth.

Semantic segmentation, by contrast, has benefited from the availability of annotated underwater datasets, though these often lack information about optical water properties. Inspired by large-scale terrestrial benchmarks such as COCO \cite{lin2014microsoft}, ADE20K \cite{zhou2017scene}, and SA-1B \cite{kirillov2023segment}, recent underwater datasets aim to provide pixel-level annotations across various object categories. The USIS10K dataset \cite{lian2024diving} is a large-scale collection of labeled images intended for salient instance segmentation in underwater environments. Other datasets such as SUIM \cite{islam2020semantic} and UIIS \cite{lian2023watermask} offer multi-class semantic annotations and cover a range of underwater scenes. However, most of these datasets focus on shallow-water conditions, and few account for variability in optical water properties or illumination conditions, despite the open question of how such variability may bias segmentation performance. 
 
The proposed ocean recipes enable the creation of evaluation datasets enriched with detailed metadata on the optical properties of the water. This supports the analysis of potential biases introduced by environmental conditions and facilitates the informed selection and robust development of semantic segmentation methods for underwater applications. 

\section{Development of Optical Ocean Recipes}
We designed the Optical Ocean Recipes to approximate the optical properties of natural seawater in a controlled tank environment, using widely available and standardized ingredients and a practical experimental setup. Our approach builds on three key principles in ocean optics: 1) Limited visibility and color distortion in underwater environments arise from the attenuation of light as it travels through water. This attenuation is described by the Lambert–Beer law (cf. to \cite{mobley1994light} for an in depth discussion):

\begin{equation}
I = I_0 e^{-(a(\lambda)+b(\lambda)) \cdot x} \qquad.
\label{eq:lambertbeer}
\end{equation}

According to this relation, light of the initial intensity $I_0$ decreases exponentially with propagation distance $x$ due to the combined effects of absorption $a(\lambda)$ and scattering $b(\lambda)$. These coefficients are intrinsic properties of the medium and depend on the wavelength $\lambda$ of the incident light. Because both absorption and scattering vary across the visible spectrum, they selectively diminish certain wavelengths more than others, leading to the characteristic color shifts in the underwater domain. Turbidity, another effect that is particularly strong in coastal regions, is primarily driven by scattering. 2) The total attenuation of seawater $c(\lambda)$ is the sum of both effects

\begin{equation}
    c(\lambda) =  a(\lambda) + b(\lambda).
\end{equation}

3) It comprises of the single attenuation spectra of all optically active constituents in the water (such as, for example, phytoplankton or colored organic dissolved matter). \cite{mobley1994light}

\begin{equation}
    c(\lambda) = \sum^i \left( a_i(\lambda) + b_i(\lambda) \right)
\end{equation}

\subsection{Color and turbidity agents} 
\label{sec:dataset_water}

As a reference for natural seawater, we use measurements by Williamson and Hollins \cite{williamson2022measured}. They acquired absorption and scattering spectra for six of the Jerlov water types \cite{jerlov1976marine}. Our approach to creating artificial seawater in a tank with the visual appearance of the reference Jerlov water type builds on mixing calibrated base additives in proper proportions. We choose three different food colorants and a scattering material as base additives. For practicality, we limit ourselves to those four ingredients, though the general approach allows to use tens or even hundreds of ingredients under certain conditions. We further address recipe composition in the subsequent section. Here, we first focus on the selection of ingredients.

\begin{figure}
    \centering
    \includegraphics[width=0.6\linewidth]{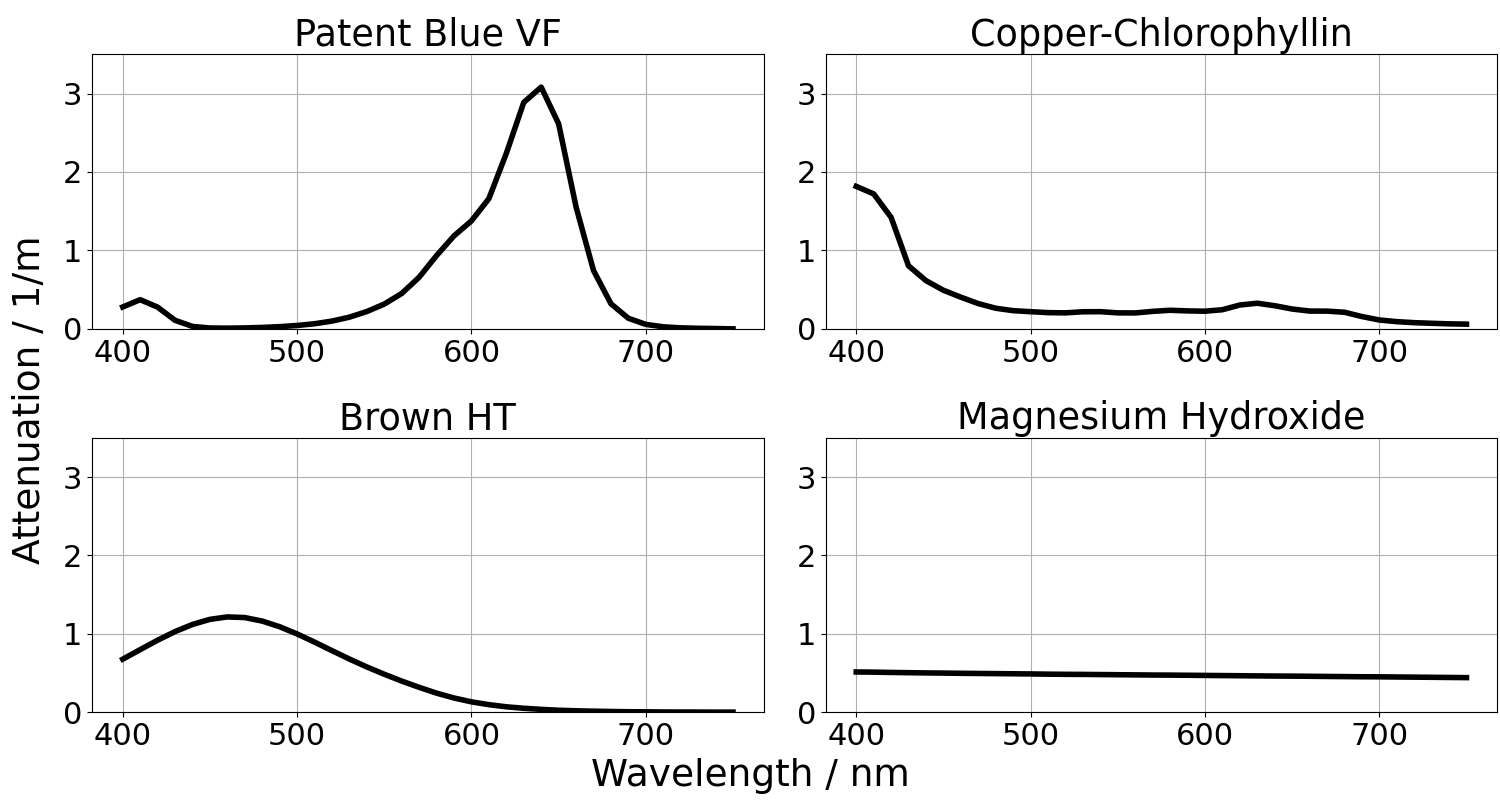}
    \caption{\textbf{Attenuation spectra of base additives.}}
    \label{fig:bases}
\end{figure}

Food colorants are a practical choice because they are typically not dangerous, not expensive and purchasable as pure standardized substance. Therefore, they can be handled with minimal chemical effort. 

Many food colorants offer a second convenient property that supports the computation of Optical Ocean Recipes: they dissolve at the molecular level and thus exhibit negligible scattering \cite{taniguchi2024light}. In contrast, scattering arises from suspensions containing particles significantly larger than water molecules or with markedly different refractive indices \cite{taniguchi2024light, kitchener2017review}. This enables controlling the absorption of artificial seawater independently from its scattering. 

The difference between molecular dissolvers and particulate scatterers becomes also visible in their attenuation spectra, which enables the separation of absorption and scattering spectra. The dissolved colorants typically show non-monotonic spectra with one or two distinct peaks, whereas particulate scatterers like magnesium hydroxide exhibit smooth, nearly linear attenuation curves across the visible range \cite{taniguchi2024light}.
We measure the attenuation spectra of all base additives and the resulting additive mixes using a transmission spectrophotometer (HITACHI U2800 UV-VIS). 
The attenuation minimum indicates where the absorption drops and thus gives the magnitude of the scattering contribution. We can fit a linear curve with the slope of magnesium hydroxide to this value, yielding the scattering spectrum. Subtracting this fit from the total attenuation spectrum gives the absorption spectrum.

The International Numbering System (INS) offers a global labeling standard for food additives. In the EU and EFTA, approved additives are labeled with matching E-numbers (e.g. E141), providing a practical reference to particular substances \cite{fao2024codex, eu_Enumbers}. 
We recommend particularly three colorants that fit the seawater color space and used those for the presented experiments. Copper-chlorophyll (E141) is a derivative of natural chlorophyll that dissolves in water. Natural chlorophyll appears in algae and phytoplankton and causes the absorption of blue and red light, resulting in green color.
Brown HT (E155) absorbs blue and green light and therefore takes account for imitating other colored dissolved organic matter which have yellow or brown color \cite{mobley1994light}. Patent Blue VF (E131) can be used to enhance the natural absorption characteristics of pure water, when the baseline absorption of the clean tank water has to be increased. 

This becomes necessary whenever the optical properties of the artificial seawater are scaled up to create the appearance of a greater scene depth. The dimensions of our tank are 2\,m × 1\,m × 0.8\,m. Natural seawater has a minor effect on image appearance on such short distances and real-world image surveys contain typically greater scene depths. Therefore, we scale the optical properties of the water and create images that mimic looking through greater distances of water by increasing the attenuation. The attenuation of the clean tank water can only be increased by adding ingredients with similar but stronger attenuation like Patent Blue VF. This technique is grounded on the Lambert–Beer law (equation \ref{eq:lambertbeer}). It states that the attenuation of light, responsible for the impact of water on the image appearance, is governed by the product of the attenuation coefficient $c(\lambda)$ and the path length $x$ which, in that case, is the distance between the scene surface and the camera sensor. Consequently, scaling the scene-camera distance or instead scaling the attenuation coefficient has the same effect.

As scattering agent we use magnesium hydroxide (MgOH$_2$). It has negligible absorption properties, suspends fast and its spectral properties are comparable
with those observed in the oceans \cite{williamson2022measured, cochenour2014suppression, mullen2011investigation}. More importantly, its phase function happens to be similar to the phase functions measured in the oceans by Petzold \cite{mullen2011investigation}.However, we observed significant precipitation (also reported by \cite{mullen2011investigation}) which requires continuous mixing and limits the working time with known scattering property in the tank. Nevertheless, we decided on MgOH$_2$ as the best fit for our purposes and discuss the handling of precipitation in detail in subsection \ref{seq:reproducability}.

\subsection{Recipes for Jerlov Water Types}
The aim of the Optical Ocean Recipes is to provide those proportions of optical additives that are required to give a specified volume of clear water the target reference absorption and scattering spectrum. Analogous to the seawater, the total attenuation of the water body in our tank $c(\lambda)$ results from the sum of attenuation spectra of its optically active additives and the clear water itself:

\begin{equation}
    c(\lambda) = c_{bl}(\lambda) + c_{gr}(\lambda) + c_{br}(\lambda) + c_\text{scat}(\lambda) + c_w(\lambda)
\end{equation}

\noindent where $c_w(\lambda)$ is the absorption spectrum of pure water \cite{pope1997absorption} and $c_\text{bl}(\lambda)$, $c_\text{gr}(\lambda)$, $c_\text{br}(\lambda)$, and $c_\text{scat}(\lambda)$ are attenuation spectra of the additives Patent Blue VF, Copper-Chlorophyllin, Brown HT, and magnesium hydroxide, respectively. In the following, we refer to the colorants as blue, green and brown for simplicity. 

We calculate the required proportions in two steps: First, we fit the absorption spectra of the blue, green, and brown base colors $a_\text{i} (\lambda)$ to the reference absorption spectrum $a^\text{ref} (\lambda)$. Second, we calculate the required amount of magnesium hydroxide needed to account for the remaining difference to the target attenuation spectrum. We suppress the dependence on $\lambda$ in further notation for better readability. Instead, bold letters will indicate higher dimensionality of the spectral components.

The target absorption $\bm{\hat{a}}$ is the sum of the absorption of base colors and tank water $\bm{a_\text{i}}$, weighted by the respective proportions $V_i$ and dissolved in the total resulting volume $V$: 

We approximate the reference absorption $\bm{a^\text{ref}}$ with a linear combination of the base absorptions $\bm{a_i}$. The concentration of base colors in the target volume $V_i/V$ weights the impact of the respective bases:

\begin{equation}
      \bm{\hat{a}} = \frac{1}{V} \sum^i \bm{a_i} V_i \qquad \text{with} \qquad i \in \{ \text{bl, gr, br, w} \}
\end{equation}

We make two simplifying assumptions regarding the clean tank water. The tank has a volume of 1600\,l, whereas the volumes of base colors can be expected to be below 0.5\,l, resulting in a concentration of colors in tank water that is below 0.3\,\textperthousand. We neglect this small part in the total volume $V$. 
Additionally, despite using a high-resolution spectrophotometer (HITACHI U2800 UV-VIS) to measure attenuation spectra, we cannot observe that the clean tank water has higher attenuation than the absorption of pure water, that is known from \cite{pope1997absorption}. Consequently, we assume for $\bm{a_\textbf{w}}$ to be $\textbf{0}$ and omit it from our calculation. 

The resulting relation is an overdetermined linear system of equations:

\begin{equation}
    \begin{bmatrix}
        \bm{a_\textbf{bl}} & \bm{a_\textbf{gr}} & \bm{a_\textbf{br}}
    \end{bmatrix} 
    \begin{pmatrix}
        w_\text{bl} \\ w_\text{gr} \\ w_\text{br} 
    \end{pmatrix}
    = \bm{\hat{a}}
    \qquad \text{with} \qquad  w_i = \frac{V_i}{V}.
\end{equation}

We estimate the weights $w_i$ using a least-squares solver, constraining them to be non-negative due to their physical interpretation as volumes. Our implementation uses the non-negative least squares (NNLS) solver introduced by \cite{bro1997fast}, as provided in SciPy \cite{2020SciPy-NMeth} version 1.14.1.

This method imposes two conditions on the selection of base colors. First, the number of base colors must be smaller than the spectral resolution of their absorption spectra, that is, fewer than the number of wavelengths for which absorption coefficients are known. This ensures that the system has more equations than unknowns, yielding a unique solution. Second, the absorption spectra of the base colors should be linearly independent, which excludes colors that are mixtures or dilutions of other base colors. 

After calculating the colors to fit the target absorption, the required proportion of the scattering agent hast to be determined to add turbidity. The base spectrum of magnesium hydroxide $\bm{b_s}$ is measured from the base suspension, obtained from suspending an initial amount $m_0$ of magnesium hydroxide powder to a volume $V_s$ of clear water. The amount required to approximate the reference attenuation spectrum $\bm{\hat{c}}$ is the initial amount scaled by the weight $w_s$

\begin{equation}
    m_s = w_s \cdot m_{s}
\end{equation}

where $w_s$ is calculated as

\begin{equation}
    w_s = \frac{\bm{c^\text{ref}} - \bm{\hat{a}}}{\bm{b_{s}}} \cdot \frac{V}{V_s}
\end{equation}

\noindent from the reference attenuation $\bm{c^\text{ref}}$, the reference absorption $\bm{\hat{a}}$, the scattering spectrum $\bm{b_{s}}$ of the base suspension, and the relation between volume of base suspension $V_s$ and target (tank) volume $V$.

\section{Experiments}
The proposed Optical Ocean Recipes Method was evaluated for its ability to generate visually realistic seawater. Recipes were computed for three Jerlov water types: oceanic types IB and II, and coastal type 3C. To assess the realism of both our recipes and previously applied additives, we compared them against the absorption and scattering spectra of the reference water types. Table \ref{tab:recipes} presents the computed recipes for those three types. We scaled the water properties by five which translates actual distance between camera and scene surface of, for example, 0.5\,m to a visual appearance of 2.5\,m.

\begin{table}[ht!]
\centering
\renewcommand{\arraystretch}{1.3} 
\setlength{\tabcolsep}{6pt} 
\begin{tabular}{lccc}
\toprule
Ingredient & Type IB & Type II & Type 3C \\ 
\midrule
\rowcolor{gray!20} Patent Blue VF & 221 ml & 221 ml & 234 ml \\ 
Co – Chlorophyllin & 25 ml & 39 ml & 115 ml \\ 
\rowcolor{gray!20} Brown HT & 15 ml & 33 ml & 163 ml \\ 
Mg(OH)$_2$ & 4.4 g & 4.4 g & 21.0 g \\ 
\bottomrule
\end{tabular}
\vspace{0.5em}
\caption{Recipes to generate water appearance for Jerlov water types IB, II, and 3C as observed by \cite{williamson2022measured}.}
\label{tab:recipes}
\end{table}

\begin{figure}[ht!]
    \centering \includegraphics[width =1\textwidth]{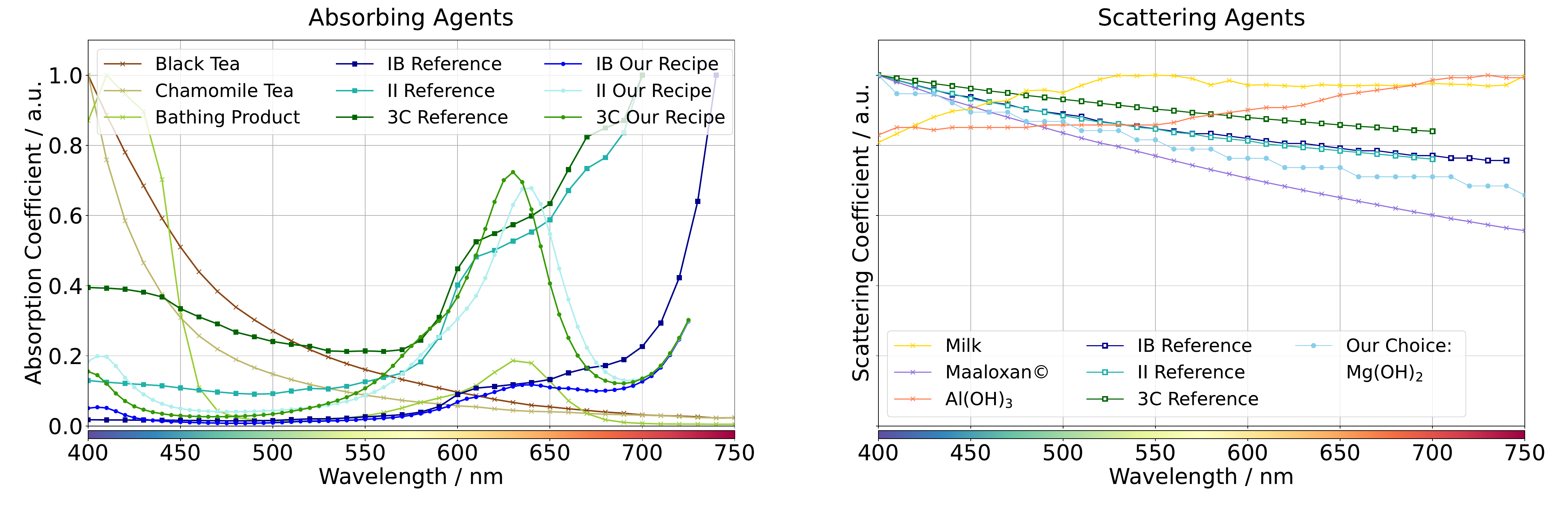}
    \caption{\textbf{Comparison of absorption spectra (left) and scattering spectra (right)} of reference waters measured by from Williamson/Hollins \cite{williamson2022measured}, our recipes and optical additives from previous work. All spectra are normalized to the maximum such that their shapes are independent of concentration. The color bar illustrates the color of light corresponding to the respective wavelength.}
    \label{fig:jerlov_spectra}
\end{figure}

\begin{figure}[ht!]
    \centering
    \includegraphics[width=\linewidth]{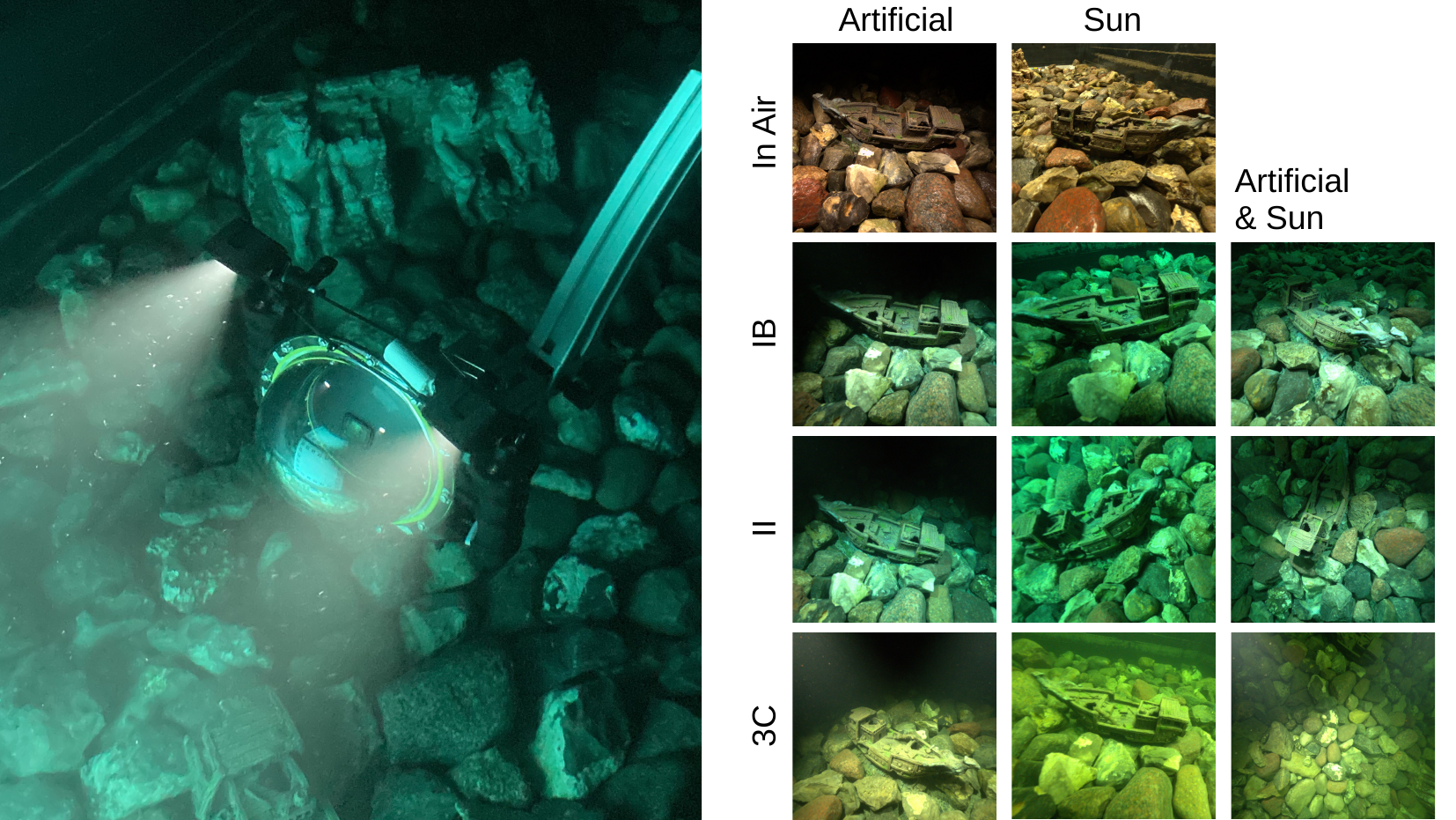}
    \caption{\textbf{Camera Set up (left) and example images} in three Jerlov water types and three illumination settings. }
    \label{fig:cam_set_up}
\end{figure}

Qualitative results are presented in the figures \ref{fig:Water_SetUp} and \ref{fig:cam_set_up}. Note, for example, how the two example images in figure \ref{fig:Water_SetUp} visualize that the difference in attenuation spectra between Jerlov 3C waters and Jerlov IB waters does not only change the color of the water. Also the visibility diminishes from the IB to the 3C example.
The quantitative comparison of our results to previously used additives and the reference waters are illustrated in figure \ref{fig:jerlov_spectra}.

\subsection{Absorption}
We begin by analyzing the absorption spectra (Figure \ref{fig:jerlov_spectra}, left) of our recipes, the real reference waters, and additives used in previous studies, specifically black tea, chamomile tea, and a green-colored children's bath product. Prior to the experiment, we hypothesized that black tea might help simulate the brownish tint observed in some coastal waters. However, our coastal water type exhibits a predominantly green hue. To provide a more fair comparison, we also tested chamomile tea and the green bath product. We found that the teas and bath product exhibited absorption spectra that differ significantly from those of the reference seawater types.

Both teas show strong absorption in the blue band, which decreases markedly towards the red band. As a result, black tea appears brown to reddish-brown, while chamomile tea, which already drops in absorption in the green band, produces a green-brown hue. Compared to seawater, both teas exhibit significantly lower absorption in the red band. The green bath color absorbs red light more effectively, but its blue absorption is too strong, resulting in an overly green appearance. This analysis shows that the dominant absorption effect in natural seawater, the attenuation of red light, cannot be accurately reproduced using the tested teas or the green bath color.

Our approach using multiple different base colors offers more flexibility to approximate a reference spectrum. And we can observe indead a closer fit to the target sea water spectra especially in the absorption capacity of red light. 
We observe the main deviations between our results and the sea water references in the absorption decrease above 640\,nm and a lower absorption than the Jerlov 3C and, to a minor extent, the Jerlov II reference in the blue and green band.
We expect closer fits to the target spectra by employing more base colors. The trade-off between practicability and higher precision with more base colors is a goal of further investigations. Additionally, we observed that tea, being a natural product, does not have constant color properties. This impedes controlled dosage and estimation of color properties of the created images in tea-colored water. 

\subsection{Scattering}
\label{seq:scattering}
We also assessed four scattering agents. Maaloxan\textsuperscript{\textcopyright} (or Maalox\textsuperscript{\textcopyright} is a substance known for a volume scattering function comparable to sea water and was used in earlier studies \cite{mullen2011investigation}. It is a stomach medicine that is marketed under various names, therefore it is challenging to determine whether the optical properties remain the same. Magnesium hydroxide and aluminium hydroxide are ingredients of Maaloxan\textsuperscript{\textcopyright} that are available as pure substance and are thus more easy to control. We also found milk as scattering agent in the literature \cite{codevilla2004achieving, garcia2011detection} and included it in our analysis. 
The results (figure \ref{fig:jerlov_spectra} left graph) show that magnesium hydroxide is the closest to the reference spectra. 
From the four tested scattering agents Maaloxan\textsuperscript{\textcopyright} shows a shape that decreases with longer wavelength too, but its slope is slightly steeper. Aluminium hydroxide and milk have in turn an inverted spectrum resulting in more scattering of red light and less of blue as compared to seawater. 

\subsection{Reproducibility and Interpretability}
\label{seq:reproducability}

The food colorants we used are highly concentrated powders, which poses challenges for reproducibility
The required amounts of powder to dye the tank are on the order of a few micrograms, which approaches the lower limit of many precision scales. Measuring the colorant powder is therefore impractical and prone to large uncertainties. Instead, we dissolve an approximate quantity of powder in water. We then measure the attenuation spectrum of this base solution and use it to calculate the volume required for each recipe.
As a result, each base solution exhibits slight variations in its attenuation magnitude.
Consequently, the recipes for the same target attenuation spectrum may vary depending on the specific base solution used.
To be able to redo the same recipe with different base solutions, we report our recipes with respect to \textit{virtual} standard solutions. We consider a standard solution to have an attenuation spectrum that integrates to 1 mm$^{-1}$ over the visual spectrum (between 400 nm and 700
nm). Each new base solution has a different scaling factor to the standard solution that has to be determined with an attenuation measurement device before calculating a recipe for an experiment. 

Nevertheless, the recipes normalized to the standard solutions are better comparable, reproducible and interpretable. We reported those in table \ref{tab:recipes}. Higher proportions of, for example, green color in a recipe result in actual more green appearance in the tank. You can observe in the recipes calculated for the experiments reported in this work, that blue is required in the largest quantity and in the same amount over all types. This effect is likely due to upscaling the attenuation of the entire water body, including its natural blue coloration, to enhance the scene depth in tank images, as discussed in Section~\ref{sec:dataset_water}. We can also observe how our recipes create the different appearances between the two images in figure \ref{fig:Water_SetUp} a). The required proportions of green, brown and magnesium hydroxide increase from water IB to water 3C approximately by factor five to ten. This differences create the visible color shift and reduction in visibility. 
This correspondences can not be observed, when the base solutions are not normalized to the same attenuation. 

\subsection{Uncertainties}
\label{seq:uncertainties}
While it is theoretically possible to compute a recipe that mimics a target spectrum, in practice, the input base spectra already carry uncertainties. 
Additional sources of uncertainty arise during the mixing process, which involves calculating, weighing, measuring, and dispensing the required amounts of base ingredients. Further variability is introduced in the control process, where the optical properties of the prepared tank water are measured and evaluated. These uncertainties stem primarily from the limited resolution of the measurement instruments. However, the precipitation of magnesium hydroxide also makes a significant contribution.

To begin with, consider the uncertainties introduced by the measurement instruments. The mixing process involves the spectrophotometer to measure the spectra of base solutions, a precision scale to weigh the required amount of magnesium-hydroxide, a scientific measuring cup with 5 ml resolution and a standard folding ruler to measure water depth in the tank for an estimate of its water volume. Note that the measuring cup introduces uncertainty at two stages: first, when the base solutions are diluted with clear water to bring their absorbance within the spectrophotometer’s working range (a maximum absorbance of 3 in our case); and second, when the final volume of each component is measured.

\begin{figure}[ht!]
    \centering
    \includegraphics[width=0.8\linewidth]{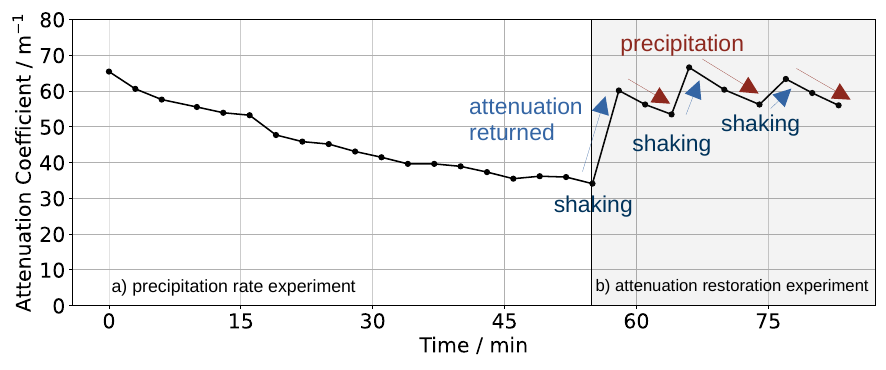}
    \caption{\textbf{Attenuation of a magnesium hydroxide suspension over time}. Precipitation rate was initially monitored over 55 minutes (a). At 55, 64, and 74 minutes, the cuvette with the suspension was removed from the spectrophotometer, shaken, and reinserted (b). Each shaking restored attenuation to near-initial levels.}
    \label{fig:precipitation}
\end{figure}

In addition to these instrumental limitations, the precipitation of magnesium hydroxide has to be considered. To this end, we performed a time series measurement of the attenuation rate of a magnesium hydroxide suspension. We observed an attenuation loss of 7.3\.\% over the first 3 minutes and 15.6\.\% over the first 12 minutes (see figure \ref{fig:precipitation}). 

After 55 minutes, we tested how much of the initial attenuation could be restored by shaking the probe, simulating the effect of tank water stirring. Each shaking event restored attenuation to near-initial values.
While a single time series is insufficient to draw general conclusions about the precipitation dynamics of magnesium hydroxide, this experiment informed both our handling of the material and our estimation of measurement uncertainty due to precipitation. Specifically, each probe was shaken prior to measurement to minimize attenuation loss.
And to ensure consistency between the measured optical properties and the appearance of images taken in the tank, we stirred the water before each imaging session and before every control attenuation measurement.

Measurement uncertainty due to precipitation contributes to the overall uncertainty when measuring the base spectrum of magnesium hydroxide. Since the measurement process takes one to two minutes, we estimate an attenuation decay of up to 7\.\% during this period. This implies that the true attenuation may be up to 7\.\% higher than the measured spectra. 

\begin{table}[ht!]
\centering
\renewcommand{\arraystretch}{1.3} 
\setlength{\tabcolsep}{6pt}
\resizebox{0.7\columnwidth}{!}{
\begin{tabular}{l|l}
\textbf{Source of Uncertainty}    & \textbf{Tolerance Range}       \\ \hline
Measuring cup resolution          & $\pm$ 2.5 ml                    \\
Scale reproducibility             & $\pm$ 0.02 g                    \\
Scale linearity                   & $\pm$ 0.03 g                    \\
Folding ruler                    & $\pm$ 0.5 mm                    \\
Magnesium hydroxide precipitation & $-$7\% attenuation (after 3 minutes)      \\
\multirow{3}{*}{Spectrophotometer tolerance} 
                                  & $\pm$ 0.002 (0–0.5)             \\
                                  & $\pm$ 0.004 (0.5–1.0)           \\
                                  & $\pm$ 0.008 (1.0–2.0)           \\
\end{tabular}
}
\vspace{0.5em}
\caption{Uncertainties in the process of creating and testing the ocean recipes.}
\label{tab:uncertainties}
\end{table}

We numerically evaluate the effects of spectrophotometer uncertainty, tank volume uncertainty, measuring cup uncertainty, and precipitation uncertainty of magnesium hydroxide on the resulting recipes by repeating the recipe calculation across the accumulated upper and lower bounds of each instrument’s measurement tolerance. The maximum resulting variation remains below 7\,\%. In absolute terms, this corresponds to variations between 0.9\,ml and 15.5\,ml for liquid volumes, and between 0.2\,g and 1.3\,g for the required amount of magnesium hydroxide.
In some cases, this level of uncertainty is already lower than the resolution of the measuring cup and scale used in the subsequent step to prepare the actual mixtures. Nevertheless, these instruments contribute additional uncertainty, as specified in Table\,\ref{tab:uncertainties}, which contributes to the final quantities dispensed into the tank.
  
The reported resulting spectra were measured prior to dispensing the ingredient mixture into the tank. This approach minimizes the larger uncertainties associated with measuring liquids of low absorbance, where the influence of confounding particles or impurities, originating from transport containers or the measuring cuvette, can be disproportionately high. These confounding effects are particularly difficult to quantify, as they may not be visually detectable. Finally, the resulting spectra of the ingredient mixtures exhibit the same level of uncertainty, arising from the spectrophotometer and the measuring cup, as the spectra of the base solutions.

\section{Image Dataset \& Vision Applications}
Having obtained recipes for three Jerlov water types, we create a demonstration image dataset and illustrate how the images generated with the proposed method enable detailed evaluation of underwater feature matching and instance segmentation. It is important to note that our evaluations are conducted on relatively small datasets and are intended for demonstration purposes. For more robust assessments of algorithm performance under specific optical underwater conditions, our ocean recipes framework can be used to generate larger and more comprehensive image collections.

\subsection{Demonstration Dataset}

Images were taken with a GoPro 9 with ultra-wide lens in time-lapse mode. To account for a large dynamic range we used auto-exposure mode and additionally stored images in GoPro RAW format. The camera was placed in a housing with dome port. Radial distortion due to the wide lens was corrected by the fisheye camera model implemented in Colmap. 
The image sets contain still photos with random pose variations and a wide range of perspectives. The same capture procedure was repeated in the three different water types as well as without water. Note that the set of poses differs for each water type, as all images were captured hand-held.

We selected three illumination settings for our experiments. To simulate imaging conditions in deep ocean regions, where sunlight is minimal or absent, we positioned lamps on both the left and right sides of the camera, angled toward its optical axis, and captured images in an otherwise darkened laboratory environment. In contrast, shallower regions receive some sunlight, which we replicated in the lab using three ceiling-mounted lamps equipped with wide diffusers to produce a more diffuse, ambient light. The third illumination setting combined artificial lighting with simulated sunlight to represent intermediate lighting conditions.

To generate ground-truth poses for image matching evaluation, we suggest leveraging the advantage that the tank has over the real ocean and drain the water from the scene. It is then possible to robustly conduct a sparse reconstruction of the acquired images. We applied COLMAP \cite{schoenberger2016sfm} for this task. Afterwards, we register each of the underwater images we captured to the in-air reconstruction of the tank scene and obtain the ground-truth pose. \footnote{Note that the term ground-truth refers to data that is considered objectively correct because it is obtained by direct measurements. Reference data that is obtained by other algorithms, like our reference poses, is more precisely named pseudo ground truth. However, for better readability, we omit the prefix pseudo here.}

\subsection{Image Matching}  
We exemplify two evaluation protocols. The first compares image matching pipelines on our dataset. The second reveals the most critical factors for the matching performance on our dataset. 
For demonstration, we selected two popular pipeline variants with a state-of-the-art evaluation protocol: one using feature detection and description from
SIFT \cite{lowe1999object}, and another using DISK
\cite{tyszkiewicz2020disk}, comparing the performance of an
engineered feature detector against a learned one. Tentative matches
are predicted using LightGlue \cite{lindenberger2023lightglue}, which incorporates global context by modifying local embeddings with a graph neural network \cite{lindenberger2023lightglue}. The matches are then input to MAGSAC for fundamental matrix estimation \cite{barath2019magsac}, a RANSAC variant with modifications tuned for epipolar geometry estimation. The two pipelines are evaluated using pose accuracy, a key
downstream task that captures any failure in the pipeline. We follow
the evaluation protocol of the \emph{Image Matching Challenge}
\cite{Jin-IJCV-2021}. 

Scale ambiguity in the pseudo ground truth reconstruction is
resolved using annotations, allowing us to compute absolute position
error. We measure mean Average Accuracy (mAA) by computing the
empirical cumulative distribution of both errors up to given
thresholds, recognizing that sufficiently large errors correspond to
random poses. For example, an error of 30 degrees is no better than
180 degrees — both indicate failure. By computing the cumulative
distribution function (CDF) of pose errors, we assign greater weight
to methods that achieve higher accuracy at lower error thresholds,
rather than relying on a single threshold-based metric.

The resulting mean average accuracies are summarized in Table \ref{tab:image_matching_results}. Overall, the SIFT pipeline outperforms the DISK pipeline in all but one subset. Interestingly, the image matching performance on the in-air datasets is not substantially higher than that observed in the clear to mildly turbid underwater conditions we created. Furthermore, the observed performance variations do not show a clear correlation with water type, lighting setup, mean image overlap, or mean scene depth. To better understand the factors influencing performance, we conduct a feature importance analysis. The controlled tank environment provides access to detailed metadata that support investigation of these factors.

\begin{table*}[ht!]
\centering
\begin{tabular}{l|cccccc|cc}
\toprule
\textbf{Data Subset}                        & \textbf{Images}             & \textbf{Pairs}        & \textbf{$\bar{z}$\,/\,m}      & \textbf{$\sigma_z$\,/\,m}   & \textbf{$\bar{\mathcal{O}}$/\,\%}      & \textbf{$\sigma_{\mathcal{O}}$\,/\,\%}  & \textbf{DISK}        & \textbf{SIFT}                \\ \hline
\cellcolor[HTML]{FFFFFF}In Air - Sun        & 75                          & \cellcolor[HTML]{FFFFFF}46  & \cellcolor[HTML]{FFFFFF} 0.62 & \cellcolor[HTML]{FFFFFF} 0.17 & \cellcolor[HTML]{FFFFFF} 0.38 & \cellcolor[HTML]{FFFFFF} 0.11 & \cellcolor[HTML]{FFFFFF} 0.39 & \cellcolor[HTML]{FFFFFF} 0.55 \\
\cellcolor[HTML]{FFFFFF}In Air - Artificial & 21                          & \cellcolor[HTML]{FFFFFF}124 & \cellcolor[HTML]{FFFFFF} 0.53 & \cellcolor[HTML]{FFFFFF} 0.20& \cellcolor[HTML]{FFFFFF} 0.27 & \cellcolor[HTML]{FFFFFF} 0.06 & \cellcolor[HTML]{FFFFFF} 0.64 & \cellcolor[HTML]{FFFFFF}  0.86\\
\cellcolor[HTML]{EFEFEF}IB - Sun            & \cellcolor[HTML]{EFEFEF}85  & \cellcolor[HTML]{EFEFEF}776 & \cellcolor[HTML]{EFEFEF} 0.46 & \cellcolor[HTML]{EFEFEF} 0.10& \cellcolor[HTML]{EFEFEF} 0.34 & \cellcolor[HTML]{EFEFEF} 0.11 & \cellcolor[HTML]{EFEFEF} 0.39 & \cellcolor[HTML]{EFEFEF} 0.77 \\
\cellcolor[HTML]{EFEFEF}IB - Artificial     & \cellcolor[HTML]{EFEFEF}101 & \cellcolor[HTML]{EFEFEF}45  & \cellcolor[HTML]{EFEFEF} 0.43 & \cellcolor[HTML]{EFEFEF} 0.12& \cellcolor[HTML]{EFEFEF} 0.25& \cellcolor[HTML]{EFEFEF} 0.04& \cellcolor[HTML]{EFEFEF} 0.85 & \cellcolor[HTML]{EFEFEF}  0.88 \\
\cellcolor[HTML]{EFEFEF}IB - Sun+Artificial & \cellcolor[HTML]{EFEFEF}79  & \cellcolor[HTML]{EFEFEF}270 & \cellcolor[HTML]{EFEFEF} 0.42& \cellcolor[HTML]{EFEFEF} 0.12& \cellcolor[HTML]{EFEFEF} 0.31& \cellcolor[HTML]{EFEFEF} 0.11 & \cellcolor[HTML]{EFEFEF} 0.58 & \cellcolor[HTML]{EFEFEF} 0.72 \\
\cellcolor[HTML]{FFFFFF}II - Sun            & 25                          & \cellcolor[HTML]{FFFFFF}115 & \cellcolor[HTML]{FFFFFF} 0.53& \cellcolor[HTML]{FFFFFF} 0.12& \cellcolor[HTML]{FFFFFF} 0.41& \cellcolor[HTML]{FFFFFF} 0.15& \cellcolor[HTML]{FFFFFF} 0.77 & \cellcolor[HTML]{FFFFFF}  0.92 \\
\cellcolor[HTML]{FFFFFF}II - Artificial     & 57                          & \cellcolor[HTML]{FFFFFF}44  & \cellcolor[HTML]{FFFFFF} 0.42& \cellcolor[HTML]{FFFFFF} 0.09& \cellcolor[HTML]{FFFFFF} 0.25& \cellcolor[HTML]{FFFFFF} 0.05& \cellcolor[HTML]{FFFFFF} 0.54 & \cellcolor[HTML]{FFFFFF} 0.93 \\
\cellcolor[HTML]{FFFFFF}II - Sun+Artificial & 80                          & \cellcolor[HTML]{FFFFFF}271 & \cellcolor[HTML]{FFFFFF} 0.43& \cellcolor[HTML]{FFFFFF} 0.14& \cellcolor[HTML]{FFFFFF} 0.34& \cellcolor[HTML]{FFFFFF} 0.12& \cellcolor[HTML]{FFFFFF} 0.67 & \cellcolor[HTML]{FFFFFF} 0.55 \\
\cellcolor[HTML]{EFEFEF}3C - Sun            & \cellcolor[HTML]{EFEFEF}21  & \cellcolor[HTML]{EFEFEF}32  & \cellcolor[HTML]{EFEFEF} 0.48& \cellcolor[HTML]{EFEFEF} 0.01& \cellcolor[HTML]{EFEFEF} 0.33& \cellcolor[HTML]{EFEFEF}0.01& \cellcolor[HTML]{EFEFEF} 0.49 & \cellcolor[HTML]{EFEFEF} 0.88 \\
\cellcolor[HTML]{EFEFEF}3C - Artificial     & \cellcolor[HTML]{EFEFEF}110 & \cellcolor[HTML]{EFEFEF}53  & \cellcolor[HTML]{EFEFEF} 0.34& \cellcolor[HTML]{EFEFEF} 0.09 & \cellcolor[HTML]{EFEFEF} 0.25 & \cellcolor[HTML]{EFEFEF} 0.05 & \cellcolor[HTML]{EFEFEF} 0.67 & \cellcolor[HTML]{EFEFEF} 0.90 \\
\cellcolor[HTML]{EFEFEF}3C - Sun+Artificial & \cellcolor[HTML]{EFEFEF}36  & \cellcolor[HTML]{EFEFEF}11  & \cellcolor[HTML]{EFEFEF} 0.40& \cellcolor[HTML]{EFEFEF} 0.09& \cellcolor[HTML]{EFEFEF} 0.27& \cellcolor[HTML]{EFEFEF} 0.06& \cellcolor[HTML]{EFEFEF} 0.92 & \cellcolor[HTML]{EFEFEF}  0.94 \\
\bottomrule
\end{tabular}
\vspace{0.5em}
\caption{Mean average accuracy of feature matching and dataset statistics of the demonstration datasets. From right to left the columns are: mean scene depth $\bar{z}$ and its variance $\sigma_z$, the mean percentage image overlap $\mathcal{O}$ and its variance $ \sigma_{\mathcal{O}}$ and the mAA of the SIFT and the DISK pipeline.}
\label{tab:image_matching_results}
\end{table*}

The influence of water type, lighting, and camera pose on image matching performance is further investigated, leveraging the tank’s ability to provide reference camera poses and controlled water conditions. We demonstrate how this ground-truth data enables a feature importance analysis across the entire dataset, offering deeper insight into the factors that affect image matching performance.

\begin{figure}
    \centering
    \includegraphics[width=\linewidth]{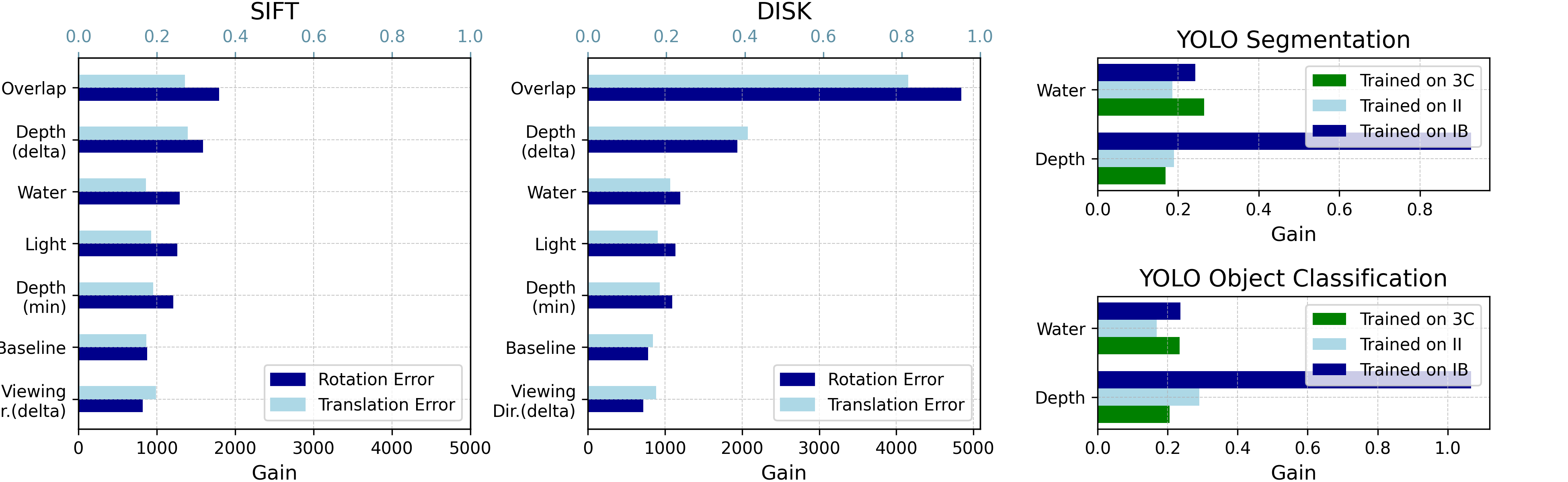}
    \caption{\textbf{Importance of Water, Light and Camera Properties} on Feature Matching or Segmentation Performance on the Demonstration Dataset containing images from clear to mildly turbid sea waters.}
    \label{fig:feature_importance}
\end{figure}

We train a regression model to capture the relationship between pose estimation error, computed from correspondences in an image pair, and a set of features expected to influence that error. The selected features include: the distance between camera centers, the angular difference between viewing directions, the smaller of the two median scene depths, the absolute difference between the median depths of both images, and the image overlap. Additionally, we include the two categorical features: water type and lighting setup. For the regression task, we employ the XGBoost gradient tree boosting framework \cite{chen2016xgboost}, chosen for its ability to model non-linear relationships and its demonstrated robustness on small (as few as 100–200 samples) and imbalanced datasets \cite{velarde2023evaluating, li2021credit, grinsztajn2022tree, bentejac2021comparative}. To encode the categorical features, we use K-fold Bayesian Target Encoding, which mitigates category imbalance by assigning higher weights to underrepresented classes. Finally, we jointly tune the regression model and target encoding hyperparameters using cross-validation.
The regression tree models partition the feature space of the matching performance data by selecting splits, one feature at a time, that minimize the difference between predicted and actual performance. Feature importance is estimated during training by aggregating the reduction in prediction error attributed to each feature across all splits in the ensemble. The prediction error is quantified using the mean squared error (MSE) between the predicted and target values. Figure \ref{fig:feature_importance} presents the gain associated with each feature, defined as the total reduction in error attributed to that feature. This gain serves as an indicator of how much each feature contributes to improving the model’s predictive accuracy.

The feature importance analysis gives more details to the outcomes of our first experiment. \textit{Overlap} and \textit{Scale difference} have the highest influence on the feature performance, even before the water and light conditions. We can also observe that those two features have an even higher influence on the mAA in the DISK pipeline as compared to the SIFT pipeline. 

\subsection{Semantic Segmentation} \label{segmentation_robustness_exp}
In the second use case, we exemplify YOLO11n-seg \cite{yolo11_ultralytics} in the context of underwater vision. Our primary goal is to assess how well a model trained under a single, controlled water and illumination setting generalizes to different underwater conditions, with particular attention to water type and its associated visibility challenges. This targets real-world deployment scenarios, where models often encounter environments that differ significantly from their training data.

To create training datasets, we annotated five objects (toy temple, board\_1, ColorChecker\textsuperscript{\textregistered}, board\_2, and toy ship) across all three water type datasets under the Sun-and-Artificial illumination setting. Each dataset was split into a training and validation set. 
Separate YOLOv11n-seg models were then trained on the training sets corresponding to each water type. To ensure comparable pose variations across the datasets, we calculated the standard deviations of rotation (in radians), view direction (in radians), and translation (x, y, z coordinates in meters) (see Table \ref{tab:deviation_overview}).
An overview of the number of images and object instances per water type is presented in Table \ref{tab:dataset_overview}. 

\begin{table}[ht!]
\centering
\renewcommand{\arraystretch}{1.3}
\setlength{\tabcolsep}{6pt}

\begin{tabular}{lccc}
\toprule
~ & Type IB & Type II & Type 3C \\
\midrule
\rowcolor{gray!20} total images & 46 & 46 & 46 \\
toy ship & 24 & 22 & 15 \\
\rowcolor{gray!20} toy temple & 30 & 23 & 25 \\
board\_1 & 22 & 14 & 11 \\
\rowcolor{gray!20} ColorChecker\textsuperscript{\textregistered} & 19 & 15 & 15 \\
board\_2 & 18 & 10 & 17 \\

\\[-0.8em] 
\midrule
\rowcolor{gray!20} total images & 10 & 11 & 9 \\
toy ship & 4 & 4 & 5 \\
\rowcolor{gray!20} toy temple & 4 & 6 & 4 \\
board\_1 & 6 & 2 & 3 \\
\rowcolor{gray!20} ColorChecker\textsuperscript{\textregistered} & 5 & 5 & 2 \\
board\_2 & 5 & 4 & 1 \\
\bottomrule
\end{tabular}
\vspace{0.5em}
\caption{Overview of the training (first six rows) and validation sets (last 6 rows).}
\label{tab:dataset_overview}
\end{table}

\begin{table}[ht!]
\centering
\renewcommand{\arraystretch}{1.3} 
\setlength{\tabcolsep}{2pt} 
\begin{tabular}{
    >{\raggedright\arraybackslash}m{1.4cm}  
    >{\centering\arraybackslash}m{2.4cm}    
    >{\centering\arraybackslash}m{2.4cm}    
    >{\centering\arraybackslash}m{2.4cm} }   
\toprule
~ & Type IB & Type II & Type 3C \\
\midrule
\rowcolor{gray!20} Rotation Std (rad) & 0.78 & 0.53 & 0.86 \\
View Std (rad) & [0.66 0.55 0.35] & [0.39 0.55 0.31] & [0.60 0.45 0.31] \\
\rowcolor{gray!20} Translation Std (x,y,z) & [0.36 0.40 0.26] & [0.39 0.36 0.15] & [0.29 0.36 0.25] \\
\bottomrule
\end{tabular}
\vspace{0.5em}
\caption{Overview of the pose-variation of the datasets.}
\label{tab:deviation_overview}
\end{table}

An overview of the number of images and object instances per water type for the validation sets are presented in table \ref{tab:dataset_overview}.

\begin{figure}[!t]
    \centering
    \includegraphics[width=0.6\columnwidth]{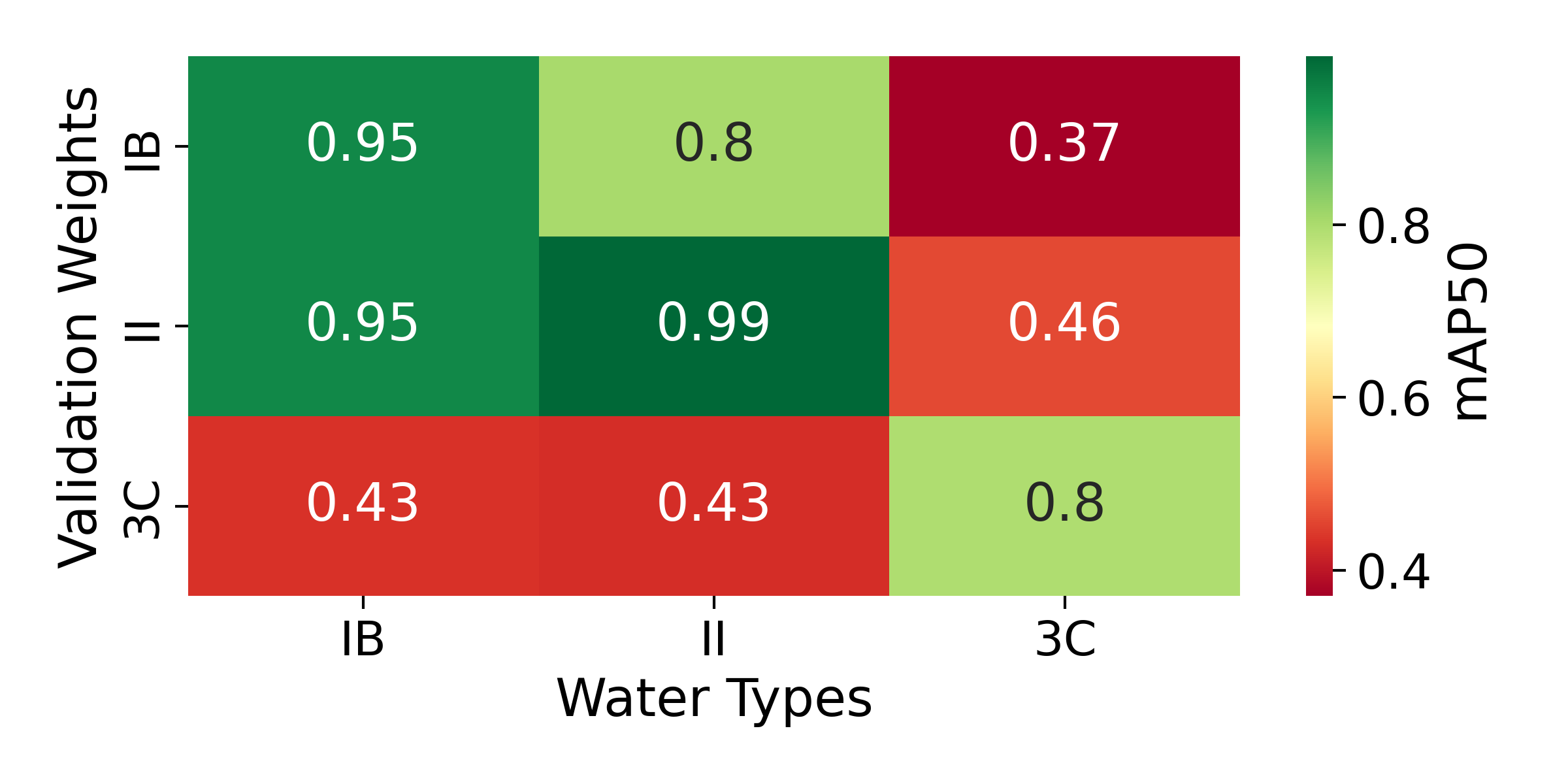}
    \caption{\textbf{Mean Average Precision (mAPmask50) across all object classes and water types} under the Sun-and-Artificial illumination setting, evaluated for models trained on each of the three water types.}
    \label{fig:heatmap_all}
\end{figure}

Segmentation performance is measured using standard metric mean Average Precision for mask predictions only considering detections where the Intersection over Union (IoU) with the ground truth is 50\,\% or greater (mAPmask50) and per-class recall. This metric reflects the model's ability to localize and correctly segment objects.
To evaluate cross-domain generalization, we suggest to validate each model trained on a specific water type on the two other datasets. The mAPmask50 scores for these cross-evaluations are shown in figure \ref{fig:heatmap_all}. As expected, the highest mAPmask50 is observed when the validation dataset matches the training domain. Notably, the performance remains relatively stable when models trained on Type IB are evaluated on Type II and vice versa  (figure \ref{fig:heatmap_all}, first and second row). However, significant performance degradation is observed when models trained on or evaluated against Type 3C are involved (Figure \ref{fig:heatmap_all}, last row), indicating a domain shift.

\begin{figure}[ht!]
    \centering \includegraphics[width =1\textwidth]{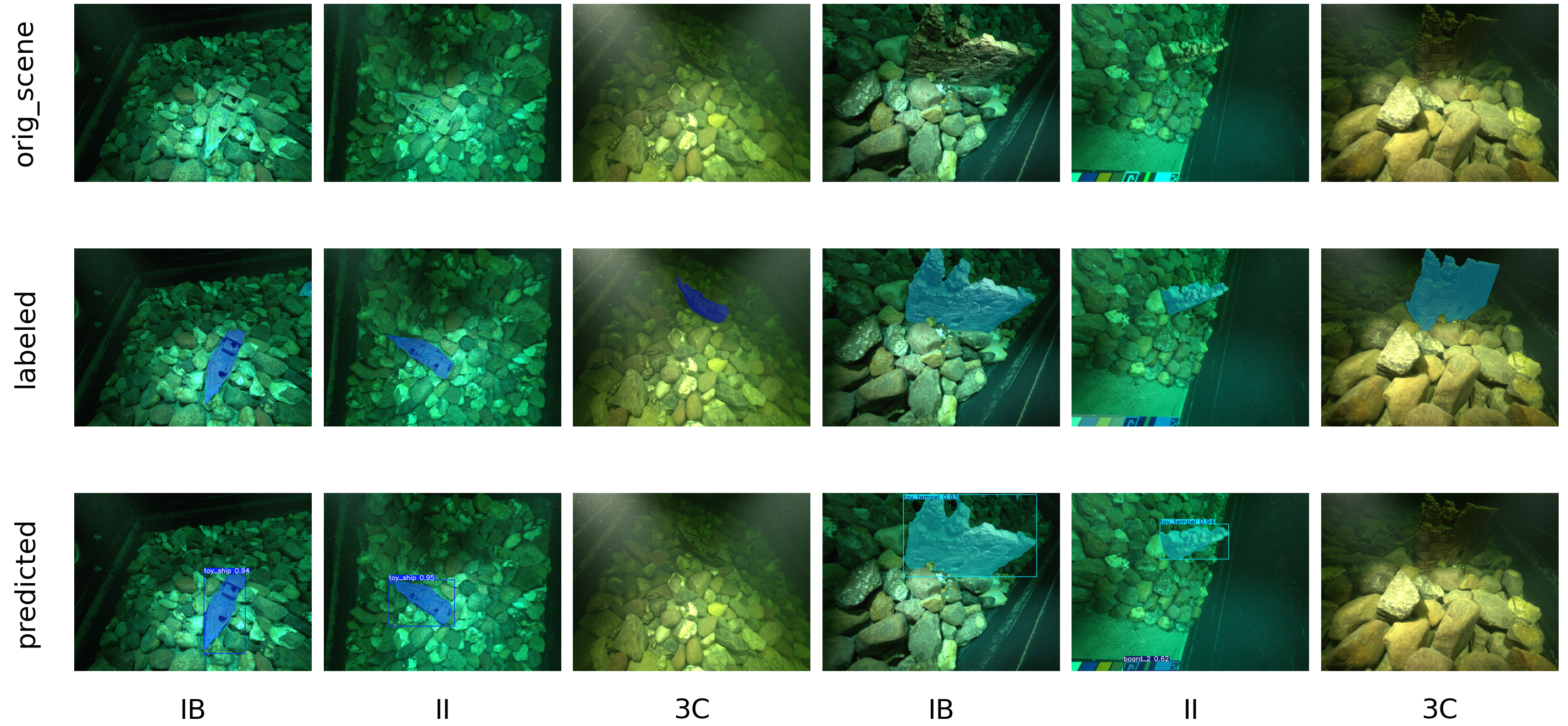}
    \caption{\textbf{Example images for each water type when predicted with the YOLO11n-seg model trained on Type IB model.} In the first scene, the toy ship is detected in Type IB and II, but missed in Type 3C. In the second scene, the toy temple is correctly segmented in Type IB and II. In Type II (second scene), board\_1 is detected but the ColorChecker\textsuperscript{\textregistered} is missed. In Type 3C, the target object is not detected at all.}
    \label{fig:seg_image_grid_Type1}
\end{figure}

\begin{figure}[ht!]
    \centering \includegraphics[width =1\textwidth]{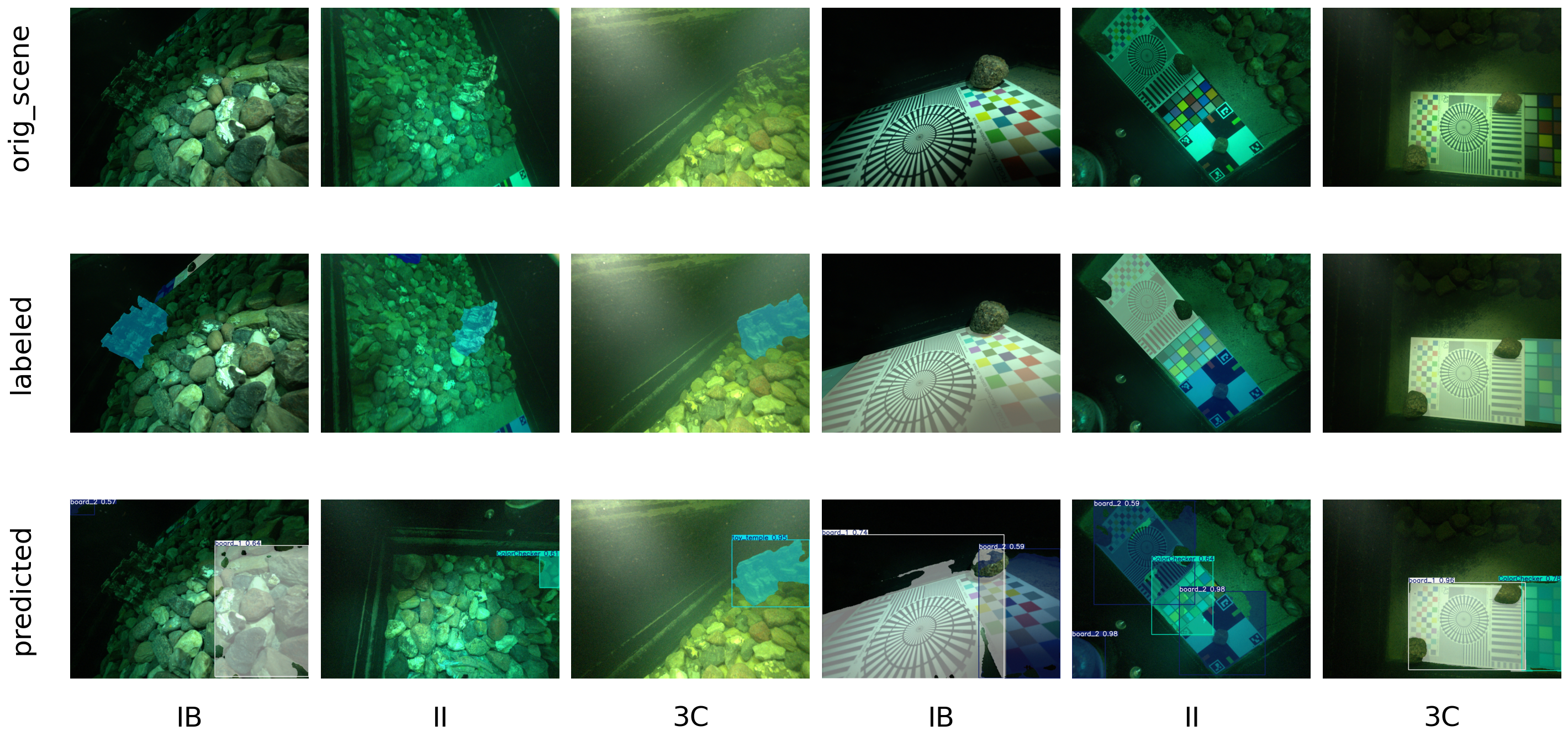}
    \caption{\textbf{Example images for each water type when predicted with the YOLO11n-seg model trained on Type 3C model.} The model correctly detects the toy temple, board\_1, and ColorChecker\textsuperscript{\textregistered} in Type 3C images. In Type IB and II (first scene), the toy temple is missed, with false positives present. In Type IB’s second scene, board\_1 is detected, but board\_2 is falsely segmented. In Type II’s second scene, board\_1 is confused with board\_2, while board\_2 and the ColorChecker® are correctly detected.}
    \label{fig:seg_image_grid_Type3}
\end{figure}

Figure \ref{fig:seg_image_grid_Type1} and Figure \ref{fig:seg_image_grid_Type3} provide qualitative insights into the generalization performance of models trained on different water types. Figure \ref{fig:seg_image_grid_Type1} shows predictions from a model trained on Type I, while Figure \ref{fig:seg_image_grid_Type3} illustrates results from a model trained on Type 3C. In general, models tend to perform well on water types visually similar to their training domain, while performance on more distinct types drops noticeably.
These trends are further supported by the quantitative evaluation shown in Figure \ref{fig:heatmap_all}. Models trained on Type I or Type II achieve high mAPmask50 when validated on Type IB and II ($\geq 0.95$), but show poor generalization to Type 3C ($\leq 0.46$). Conversely, training on Type 3C leads to better performance on its own domain ($0.80$), but significantly lower scores when evaluated on IB or II (both $0.43$). This underlines the difficulty of domain transfer between different underwater environments and emphasizes the need for either domain adaptation of segmentation algorithms or broader training datasets. We propose our optical ocean recipes to facilitate data generation for both of these purposes. 

\section{Conclusion}
\label{sec:conclsion}

We propose a novel method to generate visually realistic ocean water in a tank. It allows to compute recipes for multiple base colorants and scattering agents that create desired target appearances described by its absorption and scattering spectra. With our computed ocean recipes, we achieved a closer fit to target spectra of Jerlov water types I, IB and 3C than methods from earlier studies.  Our analysis also demonstrated that commonly used substitutes such as black tea, herbal tea, milk, or colored children's bath products exhibit significant spectral deviations from the optical properties of real seawater despite superficial visual similarities. Consequently, vision algorithms may perform inconsistently depending on factors such as scene depth or illumination conditions. 
However, our findings show that a range of realistic water types can be approximated using safe and readily available food colorants, within a reasonable margin of error. We believe that increasing the number of colorants and scattering agents could further reduce residual spectral mismatches, albeit with increased experimental complexity.
An example evaluation of image matching and semantic segmentation on image sets captured in the three water types illustrates how algorithm assessments, such as comparisons between different methods, can be linked to specific oceanic regions, thereby giving an example how more informed algorithm selection for specific use cases can be achieved. Furthermore, we demonstrate that knowledge of the optical properties of the water in evaluation images enables a systematic study of how various water-related visual disturbances affect algorithm performance. 

\section*{Acknowledgment}
This publication has been co-funded by the German Research Foundation (Deutsche Forschungsgemeinschaft, DFG) Projektnummer 396311425, through the Emmy Noether Program. We also gratefully acknowledge the financial support from the MarData Graduate School. 
Finally, the authors would like to thank the Geomar research group Marine Geosystems, specifically Bettina Domeyer and Anke Bleyer for providing access to the spectrophotometer and for their valuable advice on its use.


\clearpage
\begin{spacing}{2.0}
\bibliography{references.bib}
\end{spacing}

\end{document}